\pdfoutput=1

\documentclass[11pt]{article}

\usepackage[final]{acl}

\usepackage{times}
\usepackage{latexsym}
\usepackage{graphicx}
\usepackage{mathtools}
\usepackage{booktabs}
\usepackage{multirow}
\usepackage[T1]{fontenc}

\usepackage[utf8]{inputenc}

\usepackage{microtype}

\usepackage{inconsolata}

\title{Relay Decoding: Concatenating Large Language Models for Machine Translation}

\author{Chengpeng Fu\textsuperscript{1,2}, Xiaocheng Feng\textsuperscript{1,2}, Yichong Huang\textsuperscript{1}, Wenshuai Huo\textsuperscript{1,2},\\ {\bf Baohang Li\textsuperscript{1}, Hui Wang\textsuperscript{2}, Bin Qin\textsuperscript{1,2}, Ting Liu\textsuperscript{1,2}} \\
        \textsuperscript{1}Harbin Institution of Technology, Harbin, China \\ \textsuperscript{2}Pengcheng Laboratory, Shenzhen, China\\
        \texttt{\{cpfu,xcfeng,ychuang,baohangli,qinb,tliu\}@ir.hit.edu.cn} \\
        \texttt{\{huowsh,wangh06\}@pcl.ac.cn}\\
}

\begin{document}
\maketitle
\begin{abstract}
Leveraging large language models for machine translation has demonstrated promising results. However, it does require the large language models to possess the capability of handling both the source and target languages in machine translation. When it is challenging to find large models that support the desired languages, resorting to continuous learning methods becomes a costly endeavor. To mitigate these expenses, we propose an innovative approach called \textbf{RD} (\textbf{R}elay \textbf{D}ecoding), which entails concatenating two distinct large models that individually support the source and target languages. By incorporating a simple mapping layer to facilitate the connection between these two models and utilizing a limited amount of parallel data for training, we successfully achieve superior results in the machine translation task. Experimental results conducted on the Multi30k and WikiMatrix datasets validate the effectiveness of our proposed method.\footnote{The dataset and associated codes will be publicly available.}
\end{abstract}

\section{Introduction}

\begin{figure}[h]
\centering
\includegraphics[width=0.8\columnwidth,height=0.8\linewidth]{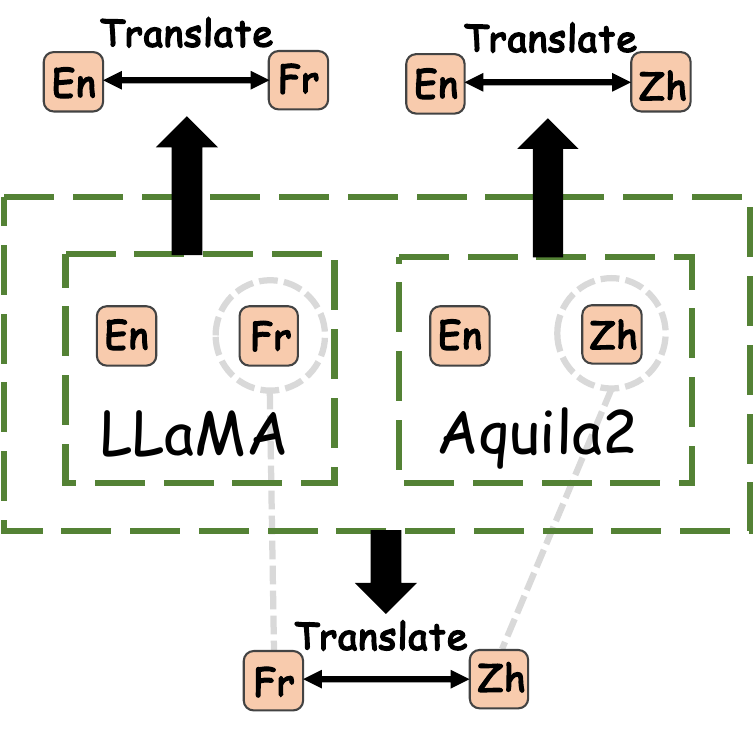}
\caption{LLaMA's supported languages include English and French while Aquila mainly support English and Chinese. Both Aquila2 and LLaMA are not proficient in handling the Chinese to French translation task individually. In such cases, we can concatenate the two models to accomplish the translation task.}
\label{fig:1}
\end{figure}

The remarkable capabilities of large language models (LLMs) with billions of parameters have been demonstrated across various tasks. Several studies have leveraged these LLMs to accomplish and enhance machine translation tasks \citep{zhang2023prompting, li-2023-practical, garcia2023unreasonable, jiao2023parrot,  lyu2023new, huang2024aligning}. Through techniques such as In-Context Learning(ICL), Chain-of-Thought(COT) and Instructions Finetuning, these LLMs have been able to achieve translation abilities comparable to state-of-the-art machine translation systems.

However, the use of LLMs for translation is still limited by the languages supported by these models. Frequently, it is challenging to find LLMs that can effectively support both the source language $L_a$ and target languages $L_b$ simultaneously, which poses a significant limitation. In such a scenario, one direct approach is to further train the existing LLM, which primarily supports one language, to incorporate the capabilities of another language\citep{cui2023efficient}. However, this requires an enormous amount of pretrained data and poses significant challenges due to the large framework of the model. Additionally, it is crucial to ensure that catastrophic forgetting does not occur, preserving the proficiency of the model in its original language.

Are there any strategies to mitigate the costly nature of continuous learning? We have observed that, in practice, it is relatively straightforward to acquire LLMs that excel exclusively in either the source or target languages. As shown in Figure \ref{fig:1}, by concatenating these specialized language models, it becomes conceivable to achieve translation without incurring the exorbitant expenses associated with continuous learning. In exceptional scenarios, when confronted with languages that lack pre-existing LLMs, a viable approach involves training a monolingual large model from scratch for the specific language. Subsequently, employing a concatenation technique enables us to accomplish machine translation, while also circumventing the issue of catastrophic forgetting in continuous learning.


Drawing on these insights, we propose a simple yet effective method \textbf{RD} (\textbf{R}elay \textbf{D}ecoding) for large model concatenation to achieve machine translation, where each large model specifically supports the source and target languages of the translation task. RD utilizes a simple mapping layer to connect two LLMs, leveraging a small portion of parallel corpora to train this mapping layer. In our experiments conducted on datasets such as Multi30k and WikiMatrix, utilizing the LLaMA and Aquila2 models, we find that our approach surpasses the method of fine-tuning with a single large model. Furthermore, we observed significant improvements of over 3 BLEU points in certain language pairs. 

\section{Approach}

\subsection{Task Description}
For a translation task from Language $L_a$ to Language $L_b$, when it is not possible to find a single large model that performs well for both languages simultaneously, we focus on finding a separate large model for each language that excels in that specific language. Let $M_a$ denote a large language model that excels in language $L_a$, and $M_b$ denote another large language model that excels in language $L_b$. RD aims to concatenate $M_a$ and $M_b$ to achieve the translation task from $L_a$ to $L_b$.

\subsection{Concatenate Method}
\begin{figure}[t]
\centering
\includegraphics[width=1\columnwidth,height=0.8\linewidth]{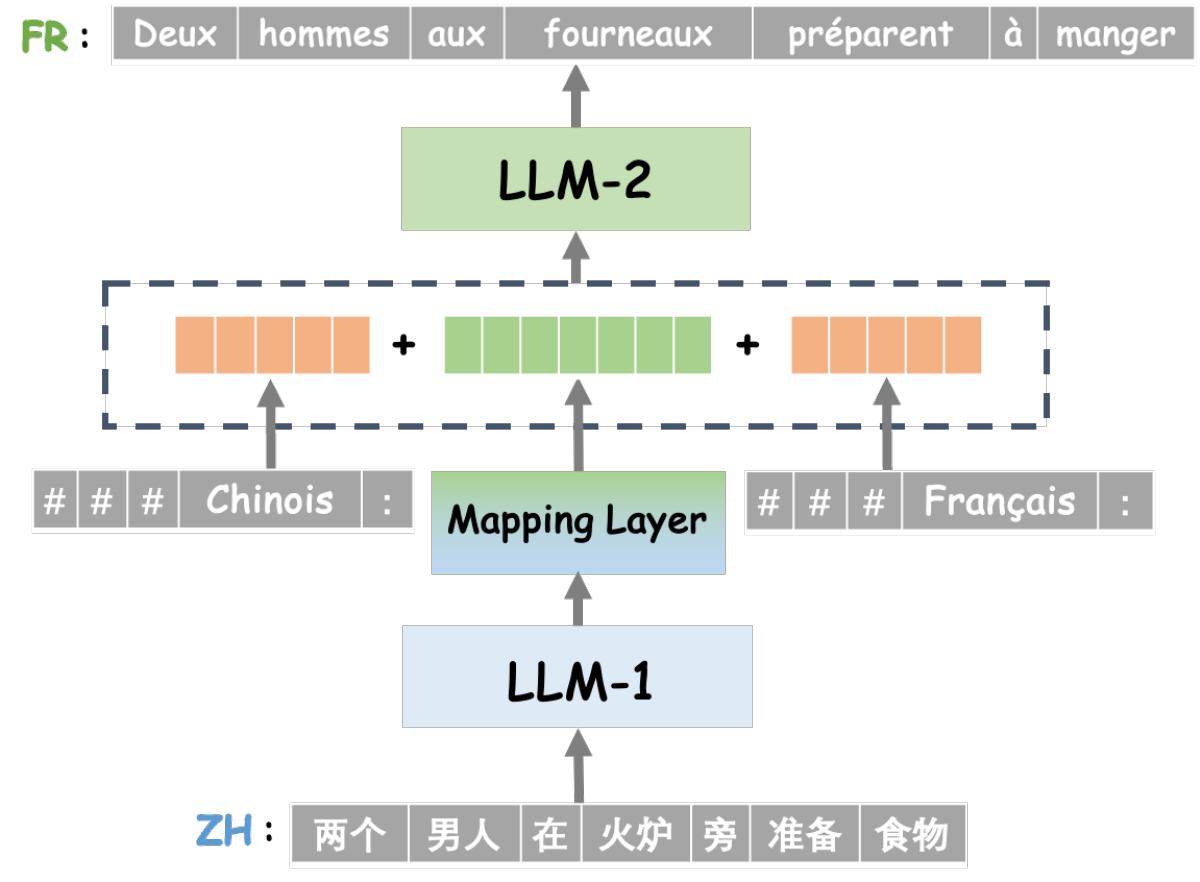}
\caption{Using Chinese-French translation as a case in point for the process of Relay Decoding.}
\label{fig:2}
\end{figure}

As illustrated in Figure \ref{fig:2}, for a given sentence $X=\{x^1, x^2,...,x^K \}$ in language $L_a$ containing $K$ tokens, we utilize $M_a$ to decode and generate its corresponding representation, denoted as $ H \in \mathcal{R}^{K*D_h}$. $D_h$ is the hidden states size of $M_a$. Subsequently, we utilize a mapping function $W_p \in \mathcal{R}^{D_h*D_e}$ to project the obtained hidden representation $H$ into the input space of $M_b$. $D_e$ is the embedding layer size of $M_b$. For the sake of simplicity and efficiency, we employ a linear layer as the mapping layer\footnote{We also attempt alternative methods of connecting the structures, which are documented in Appendix \ref{sec:appendix}.}, similar to the connection methods used in many multi-modal large models \citep{koh2023grounding, zhang2023enhanced, zhang2023transfer}.

\

Next, we introduce a prompt to facilitate better generation by $M_b$. When the source language is Chinese and the  target is English, the prompt would be as follows:
$$\#\#\# [Chinese]: X \quad \#\#\# [English]: $$

The pattern $\#\#\#[\star]$ is employed to denote the name of the specific language. In our case, we use the target language to replace these patterns.

After tokenizing the prompt and passing it through the embedding layer of $M_b$, we obtain the prompt input representation $E_p$. Finally, we concatenate the mapped representations $H$ with the prompt representations $E_p$ and feed them into $M_b$ for further decoding and generation.

\begin{table*}[]
\centering
\begin{tabular}{@{}c|cc|cc|cc@{}}
\toprule
\multirow{2}{*}{\textbf{Method}} & \multicolumn{2}{c|}{\textbf{Zh-Fr}} & \multicolumn{2}{c|}{\textbf{Zh-De}} & \multicolumn{2}{c}{\textbf{Zh-Cs}} \\ \cmidrule(l){2-7} 
 & \multicolumn{1}{c|}{\textbf{BLEU}} & \textbf{chrF} & \multicolumn{1}{c|}{\textbf{BLEU}} & \textbf{chrF} & \multicolumn{1}{c|}{\textbf{BLEU}} & \textbf{chrF} \\ \midrule
\textbf{Bilingual} & \multicolumn{1}{c|}{20.70} & 47.5 & \multicolumn{1}{c|}{10.82} & 38.3 & \multicolumn{1}{c|}{7.52} & 27.7 \\
\textbf{Aquila2} & \multicolumn{1}{c|}{19.65} & 49.0 & \multicolumn{1}{c|}{10.32} & 43.0 & \multicolumn{1}{c|}{8.75} & 36.1 \\
\textbf{LLaMA} & \multicolumn{1}{c|}{25.76} & 53.3 & \multicolumn{1}{c|}{15.00} & 49.3 & \multicolumn{1}{c|}{10.08} & 38.6 \\
\textbf{RD (Aquila2+LLaMA)} & \multicolumn{1}{c|}{\textbf{27.36}} & \textbf{55.1} & \multicolumn{1}{c|}{\textbf{17.87}} & \textbf{49.8} & \multicolumn{1}{c|}{\textbf{13.44}} & \textbf{39.1} \\ \bottomrule
\end{tabular}
\caption{The result of RD Method for Zh-Fr, Zh-De, Zh-Cs translation tasks on Multi30k dataset.  }
\label{tab:my-table}
\end{table*}

\subsection{Training Strategy}
We formulate translation task as generating target text tokens conditioned on a source text tokens and prompt prefix. The log likelihood of target sentence Y (tokenized as $\{y_1,y_2,..,y_T\}$) conditioned on its source sentence X is:
$$ l(X,Y)= \sum_{t=1}^{T} log P_\theta(y_t|H,E_p,y_1,y_2,...,y_{t-1})$$

The loss $L$ is then the negative log likelihood of all samples in a batch of N bilingual parallel pairs:

$$L=-\frac{1}{N}\sum_{i=1}^Nl(X_i,Y_i)$$

When utilizing only one LLM for translation, finetuning has been shown to yield optimal results. Therefore, in our approach, we also experiment with incorporating finetuning, which involves simultaneously adjusting the parameters of the large model during the training process. To prevent the occurrence of catastrophic forgetting, we introduced LORA\citep{hu2021lora} as a mechanism to mitigate this challenge.

\section{Experiment}
In this section, we provide a description of the datasets, experimental setup employed in our study and an in-depth analysis of the results obtained.
\subsection{Experimental Details}
\paragraph{Large Language Models} The LLMs utilized in our experiments include the Aquila2-7B model \footnote{https://github.com/FlagAI-Open/Aquila2.} and the LLaMA-7B model \citep{touvron2023llama}. In our experiments, we primarily focus on translation from Chinese to French, German, and Czech. The Aquila2 model primarily focuses on English and Chinese proficiency and performs remarkably well in tasks involving these languages. On the other hand, the LLaMA model has been pretrained on datasets encompassing twenty languages, such as English, French, German, and Czech, but its Chinese proficiency is relatively lower.
\paragraph{Datasets} The datasets used in our experiments are Multi30k dataset \citep{elliott-etal-2016-multi30k, elliott-etal-2017-findings, barrault2018findings} and WikiMatrix dataset \citep{schwenk2019wikimatrix}. Multi30k dataset contains images and their captions in four languages: English(En), French(Fr),
Germany(De), and Czech(Cs). For Chinese translation task, we have annotated a Chinese version of the Multi30k dataset\footnote{The Chinese version of multi30k will be available.}. Initially, we employ ChatGPT\footnote{https://chat.openai.com/.} to translate the English content of the dataset into Chinese. Subsequently, we conduct manual revisions to address any inaccuracies or lack of fluency in the translation. As for test set, we use Flickr2017 for Zh-Fr and Zh-De and Flickr2018 for Zh-Cs. Regarding WikiMatrix, we specifically choose the Simplified Chinese portion of the dataset. From this subset, we select the top 1000 highest-scoring pairs as our test set, while the remaining pairs are used for training.

\paragraph{Baselines}
We compared our method with the following approaches: (1) Transformer-based bilingual translation model. (2) Results of instruction fine-tuning large models, including LLaMA and Aquila2. That's an important point to note that while LLaMA may have lower proficiency in Chinese, it still has some capability in handling and generating Chinese due to the presence of a portion of Chinese data in its training set. Similarly, Aquila2 model's training corpus may also include a small amount of French, German, and Czech data. As a result, fine-tuning directly on these models can still achieve some level of performance in translation tasks for the respective languages.


\subsection{Main Results}
The experimental results on Multi30k dataset for Zh-Fr, Zh-De, and Zh-Cs are presented in Table \ref{tab:my-table}. From the table, we observe that our RD method achieves the best results. When comparing the last three rows with the first row, which represents the bilingual transformer approach, we find that utilizing large models with the same parallel corpus outperforms training from scratch. This indicates that the language alignment capability of the large models is indeed utilized during training, even though they were pretrained only on monolingual data. The results of fine-tuning large models specialized in one language (rows 2 and 3) show that these models still have some limitations in completing translation tasks. Additionally, we have also discovered that LLMs specialized in the target language tend to exhibit superior performance in translation tasks. Our concatenation method also surpasses the performance of fine-tuning with a single large model, demonstrating the need for pretraining large models on both the source and target languages to achieve better translation performance and this further validates the effectiveness of our proposed concatenation method.

\subsection{Analysis}
\paragraph{Is it necessary to finetune the LLMs during training?} Our approach involves training a mapping layer to connect two large models, but during training, we also need to consider whether to adjust the parameters of the large models. As shown in Table \ref{table2}, we test the translation performance of Zh-Fr under different finetuning conditions on Multi30K datasets and find that simultaneously finetuning the parameters of the second large model (i.e., the one specialized in the target language) yield better results. On the other hand, fine-tuning the parameters of the first large model has a less significant impact. For finetuning, we utilized the efficient finetuning method known as LORA due to its higher efficiency.

\paragraph{How much data is required to complete the training of the mapping layer?} As presented in Table \ref{table3}, we conducted Zh-Fr translation experiments using training sets of different sizes on WikiMatrix datasets. The findings reveal that on the WikiMatrix dataset, training with approximately 60,000 data points is adequate for training the mapping layer. This requirement is considerably smaller compared to the dataset size typically needed by traditional bilingual methods. Moreover, our method surpasses these methods in performance.

\begin{table}[]
\centering
\begin{tabular}{@{}c|c|c@{}}
\toprule
\textbf{Aquila2 (ZH)} & \textbf{LLaMA (FR)} & \textbf{Zh-Fr} \\ \midrule
Not Finetune & Not Finetune &  25.94\\
Not Finetune & Finetune & \textbf{27.36} \\
Finetune & Not Finetune &  23.37\\
Finetune & Finetune & 26.88 \\ \bottomrule
\end{tabular}
\caption{The BLEU scores of different finetune settings on Multi30k dataset.}
\label{table2}
\end{table}

\begin{table}[]
\centering
\begin{tabular}{@{}c|ccc@{}}
\toprule
\textbf{\#Dataset} & \textbf{2W} & \textbf{3W} & \textbf{4W} \\ \midrule
\textbf{RD} & 11.79 & 12.98 & 13.64 \\ \midrule
\textbf{\#Dataset} & \textbf{5W} & \textbf{6W} & \textbf{7W} \\ \midrule
\textbf{RD} & 14.26 & 15.44 & 15.52 \\ \bottomrule
\end{tabular}
\caption{The BLEU scores associated with varying WikiMatrix dataset sizes.}
\label{table3}
\end{table}

\section{Related Work}
\paragraph{LLMs for Machine Translation.} With the remarkable advancements of LLMs, researchers have extensively evaluated their translation capabilities using various methodologies. \citet{vilar-etal-2023-prompting, zhang2023prompting, bawden-yvon-2023-investigating} devise different prompts to facilitate translation and also examine the translation performance in various few-shot scenarios. \citet{peng-etal-2023-towards, huang2024aligning} utilize Chain-of-thought or difficulty analysis techniques to address translation challenges. In order to achieve better performance, \citet{li2023eliciting, jiao2023parrot, chen2023improving, alves-etal-2023-steering, xu2023paradigm} have explored the approach of finetuning LLMs using parallel data. All of the aforementioned methods require full support from the LLMs for the languages involved in translation. Our approach, on the other hand, is primarily designed for situations where a single large language model cannot handle all of these languages simultaneously.
\paragraph{Concatenation of LLMs.} \citet{bansal2024llm} leverages the concatenation of a smaller model and a larger model to augment the capabilities of the larger one, such as enhancing low-resource language comprehension and mathematical computation abilities. In comparison, our concatenation method is specifically tailored for machine translation tasks. Furthermore, unlike our method, they do indeed require both models to be capable of handling vocabulary from both languages involved in the translation task. 
\section{Conclusion and Future Work}
In this paper, we propose an approach that involves concatenating two LLMs, each specialized in the source and target languages, to achieve machine translation. This method circumvents the higher costs associated with continuous learning approaches. In the future, we plan to delve deeper into this concatenation method and investigate how to accomplish the connection solely with monolingual data as the finetuning approach for LLMs does not necessitate the use of bilingual data.

\section*{Limitations}
In our concatenation approach, we require a certain amount of parallel data to train the parameters of the concatenation module. Acquiring parallel data can be costly, so in the future, we plan to explore methods that rely on monolingual data and back-translation to train the parameters of the concatenation module. 



\section*{Acknowledgements}
Xiaocheng Feng is the corresponding author of this work. We thank the anonymous reviewers for their insightful comments. This work was supported by the National Key R\&D Program of China via grant No. 2021ZD0112905, National Natural Science Foundation of China (NSFC) via grant 62276078, the Key R\&D Program of Heilongjiang via grant 2022ZX01A32, the International Cooperation Project of PCL, PCL2022D01 and the Fundamental Research Funds for the Central Universities (Grant No.HIT.OCEF.2023018), Nature Scientific Foundation of Heilongjiang Province(YQ2021F006).

\bibliography{anthology,custom}

\appendix
\begin{figure*}[h]
\centering
\includegraphics[width=2\columnwidth,height=0.45\linewidth]{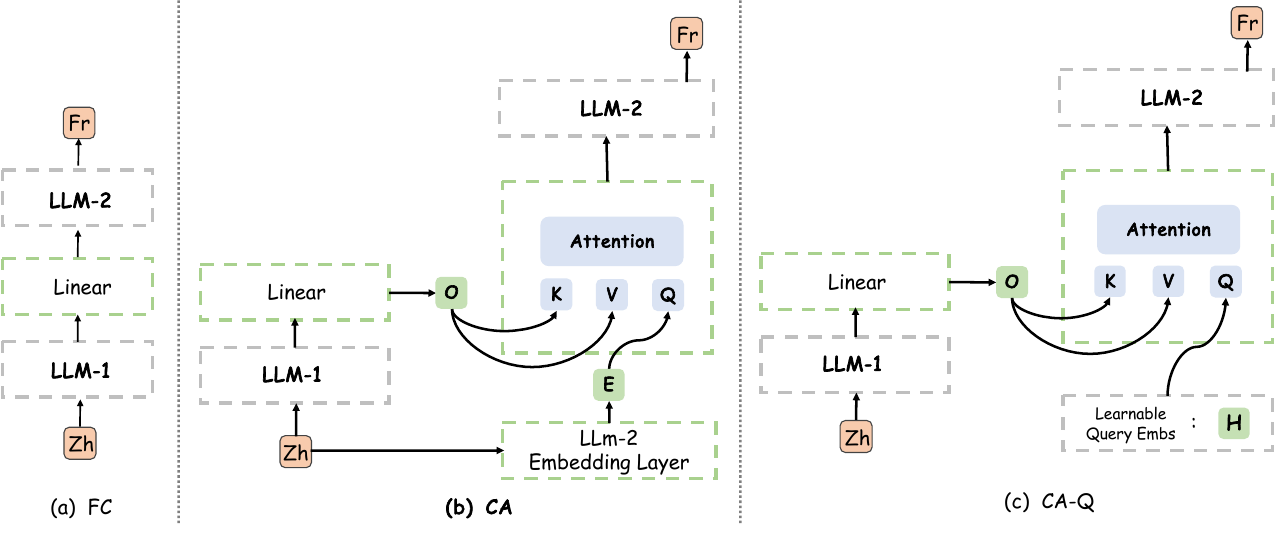}
\caption{Different mapping layers.}
\label{fig:3}
\end{figure*}\

\section{Mapping Layers}
\label{sec:appendix}

We explored three different approaches for achieving the mapping as shown in Figure\ref{fig:3}:

(1) Linear: Directly employing a linear layer, as previously mentioned, denoted by $FC$.

(2) Cross-attention: Employing a cross-attention structure with the source language input $X$, passed through the embedding layer of $M_b$, serving as the query, denoted by $CA$. 

(3) Cross-attention with query embedding: Utilizing a cross-attention structure with randomly initialized query embeddings, denoted by $CA-Q$.

We also conducted Zh-Fr translation experiments on the Multi30k dataset, and the experimental results are presented in Table \ref{table4}.

\begin{table}[t]
\centering
\begin{tabular}{@{}c|ccc@{}}
\toprule
\textbf{Mapping Method} & \textbf{FC} & \textbf{CA} & \textbf{CA-Q} \\ \midrule
\textbf{Zh-Fr} & 27.36 & 11.80 & 17.92 \\ \bottomrule
\end{tabular}
\caption{The BLEU score of different mapping method on Multi30k dataset.}
\label{table4}
\end{table}

We observe that the Linear mapping method achieved the best results on the Multi30k dataset, while the cross-attention series method yield lower results, even lower than the baseline methods. This could be attributed to the larger number of parameters introduced by these methods, which may not be effectively learned due to the relatively small scale of the Multi30k dataset. 

\section{Experiments System Settings and Evaluation Metric}
We use Adam optimizer and 2000 warm-up updates. The learning rate is 1e-5. For evaluation, we use 4-gram BLEU \citep{papineni-etal-2002-bleu} and chrF scores by multi-bleu.pl in Moses\footnote{https://github.com/moses-smt/mosesdecoder}. We train all models on NVIDIA 80GB A100 GPUs. 
\end{document}